\documentclass[sigconf,anonymous=false]{acmart}

\AtBeginDocument{%
  \providecommand\BibTeX{{%
    \normalfont B\kern-0.5em{\scshape i\kern-0.25em b}\kern-0.8em\TeX}}}

\setcopyright{acmcopyright}
\copyrightyear{2022}
\acmYear{2022}
\acmDOI{10.1145/nnnnnnn.nnnnnnn}

\acmConference[GECCO '22]{the Genetic and Evolutionary Computation Conference 2022}{July 9--13, 2022}{Boston, USA}
\acmPrice{15.00}
\acmISBN{978-x-xxxx-xxxx-x/YY/MM}

\usepackage{algorithm}
\usepackage{algpseudocode}
\begin{document}

\title{Active Learning Improves Performance on Symbolic Regression Tasks in StackGP}



\author{Nathan Haut}
\affiliation{%
  \institution{Michigan State University}
  \city{East Lansing}
  \state{Michigan}
  \country{USA}
  }
\author{Wolfgang Banzhaf}
\affiliation{%
  \institution{Michigan State University}
  \city{East Lansing}
  \state{Michigan}
  \country{USA}
  }
  \author{Bill Punch}
\affiliation{%
  \institution{Michigan State University}
  \city{East Lansing}
  \state{Michigan}
  \country{USA}
  }

\renewcommand{\shortauthors}{Haut and Banzhaf and Punch}

\begin{abstract}
In this paper we introduce an active learning method for symbolic regression using StackGP. The approach begins with a small number of data points for StackGP to model. To improve the model the system incrementally adds a data point such that the new point maximizes prediction uncertainty as measured by the model ensemble. Symbolic regression is re-run with the larger data set. This cycle continues until the system satisfies a termination criterion.  We use the Feynman AI benchmark set of equations to examine the ability of our method to find appropriate models using fewer data points. The approach was found to successfully rediscover 72 of the 100 Feynman equations using as few data points as possible, and without use of domain expertise or data translation.
\end{abstract}



\maketitle

\section{Introduction}
Symbolic regression is a typical application of genetic programming (GP) that develops mathematical models to fit data sets \cite{koza92,bnkf98}. This is a form of understandable AI since the final model can be easily presented to the end user in the form of an equation. This makes it an appealing tool for researchers attempting to understand a system of study. While many different implementations exist for symbolic regression, such as DataModeler, Eureqa, AIFeynman, etc., symbolic regression is not a solved problem \cite{datamodeler,eureqa,feynman}. Questions remain such as how much data is needed, what genetic operators to use, what representation is most effective, and what fitness function(s) should be used. 

Previous to this study, a set of 100 Feynman equations was used to compare the effectiveness of different symbolic regression implementations \cite{feynman}. This benchmark data set was used to test the ability of a machine learning (ML) system to rediscover the equations using the fewest data possible. This is a useful benchmark since all of the equations are physically meaningful. Good performance on this benchmark could indicate a ML/GP system is viable for use in scientific studies attempting to discover equations describing natural phenomena. 

Udrescu and Tegmark themselves developed an effective ML approach, AIFeynman, that is capable of solving all 100 problems in the Feynman Symbolic Regression Database \cite{data,feynman}. However, on many of the more complex equations their approach relies heavily on dimensional analysis, translation, and neural networks that take advantage of symmetries, smoothness, and separability designed specifically to solve these types of physics problems. 
While their method works extremely well in solving these problems, for cases that have to rely on neural networks, large data sets are required to rediscover the equation. As well, the dimensional analysis, translations, and assumptions about symmetry and separability required significant domain expertise, rendering this a complex approach to solving general purpose symbolic regression problems. Further, dimensional analysis requires that units both be known and recorded with the data, which may often not be the case in real-world applications.

Active learning \cite{alintro} is a machine learning strategy where an algorithm self-selects additional training data to maximally inform its own learning process. Active learning has been applied to genetic programming classification tasks where points are only labelled when the developing models encounter points that can't be classified \cite{classify}. This was found to reduce the total effort needed to label training points, since only a subset had to be labelled before finding accurate models. Active learning has also been applied to genetic programming where training sets are large by selecting sub-samples of the training data to be used. Active learning for sub-sampling was found to decrease training times to find quality binary classification models by an order of magnitude \cite{sample}.

The goal of StackGP with active learning is to create a general purpose GP system that requires no domain expertise, uses the least number of data points possible, and can guide data collection to be maximally informative. Beyond data collection for model training, the developed models could be used to design experiments to further explore the system of study. As well, the models could be used to accelerate the development process by recommending the target conditions for the system of study. An example of this could be using the developed models to design a chemical with specific target properties by recommending the conditions to produce such target properties.

\section{Related Work}

Previous work by Kotanchek et al. \cite{trustable} laid the foundation to use genetic programming for active design of experiment, where models developed by a GP system can be used to find optimal conditions in a system of study. Active design of experiment is a field closely related to active learning, since it has the goal of designing experiments that are maximally informative. They proposed using model ensembles from symbolic regression to find regions of uncertainty and exploit those regions of uncertainty to gather new data with high information content.

Active learning methods have been employed successfully to help discover biological networks 
\cite{bio}. Several different methods were explored by the authors for determining which new data points would be maximally informative. One method the authors explored was the maximum difference method in which two best-fit models are chosen and a new data point is selected where those two best-fit models have the largest difference in predictions. Another method they examined was entropy score maximization. In that method a new data point is selected that maximizes an entropy score, where the entropy can be thought of as the amount of information to be gained by gathering that data point. The entropy score $H_e$ is computed as shown in the equation below, where $M$ is the set of Boolean networks, $x_e$ is the number of network states for a given data point, and $e$ is the set of all potential data points.

\begin{equation*}
   H_e=-\sum _{x=1}^{x_e} \frac{e_x}{|M|}\log _2\frac{e_x}{|M|}
\end{equation*}

Active learning has also been applied in chemical engineering to expedite a reaction screening process by only selecting a subset of maximally informative experiments to complete rather than exhaustively performing all possible experiments \cite{chem}. This was done by training neural networks and using them to select a subset of experiments that will maximize the information gain. Maximal information gain was determined by looking at the standard deviation of an ensemble of neural networks.  

The influence of dimensional awareness on the ability for symbolic regression to rediscover 27 of the equations from the Feynman Symbolic Regression Database has recently been explored in  \cite{fitnessland}. It was found that the use of dimensional awareness can significantly reduce the computational cost of rediscovering those equations when compared to symbolic regression without dimensional awareness. The authors acknowledged that while dimensional awareness worked well, it may not be feasible in many real-world applications since it requires that all data points be labelled with units. This is often not the case, either because units were not recorded during data collection, or because the units are not known. This indicates support for alternative methods to improving symbolic regression's learning and success rates. 

\section{Methods}
 Our active learning strategy is an iterative process that trains models on data, selects an ensemble of good models, then uses the ensemble to find a new point to add to the training data that maximizes uncertainty. The 
 algorithm is summarized in Algorithm 1 and each part is described in detail in the subsections following. 

\begin{algorithm*}
\caption{Active Learning Process}\label{alg:cap}
\begin{algorithmic}
\State $Training Data \gets 3StartingPoints$                \Comment{Generate initial random training data}
\State $Models \gets RandomModels$
\Comment{Generate initial random models}
\State $Models \gets Evolve(TrainingData, Models)$                  \Comment{Train models on starting data}
\While{$Best Model Error \neq 0$}                           \Comment{While perfect model not found}
    \State $Ensemble \gets EnsembleSelect(Models)$.         \Comment{Select ensemble of models}
    \State $New Point \gets MaximizeUncertainty(Ensemble)$  \Comment{Find point that maximizes uncertainty}
    \If{$NewPoint \subset TrainingPoints$}                  \Comment{If point already selected}
        \State $NewPoint \gets MaximizeUncertainty(SubSpace(Ensemble))$\Comment{Search a subspace}
    \EndIf
    \State $Training Data \gets Append(Training Data, New Point)$ \Comment{Add new point to training data}
    \State $Models \gets Evolve(Training Data, Models)$           \Comment{Evolve new models with new data using best models to seed evolution}
\EndWhile
\end{algorithmic}
\end{algorithm*}

\subsection{Active Learning}
The goal of active learning is to strategically select new data points that are most informative to the current models. One way to identify informative points is to find 
uncertainty among the current models. The uncertainty metric $\Delta$ we apply is defined as the standard deviation of the ensemble divided by the 70 percent trimmed mean of the absolute value of ensemble responses.  

\begin{equation*}
    \Delta = \frac{\text{Std}(\text{EnsembleResponses})}{\text{TrimmedMean(Abs(}(\text{EnsembleResponses},0.3))}
\end{equation*}

The trimmed mean is used to ignore potentially asymptotic behavior that could occur in a few of the models.   Below is the step-by-step explanation of how this active learning approach works. 

\subsubsection{Initialization}
To start, 3 random data points from the region defined in \cite{feynman,data} are generated. Another 100 data points are generated as test points and are used purely for tracking the progress of the model development. They are not used to inform the model development.

An initial set of models are trained on these 3 data points. Evolution is allowed to run for up to 2 minutes running independently on 4 cores. Each run has a population size of 300 models initialized and randomly at the outset, such that the starting population consists of random models with an operator stack of 10 operators or less. An operator is any of the math operators allowed to be used during evolution plus the pop operator. StackGP's default math operators are: $+, -, *, /, Exp, Sqrt, Inverse,$ and $Squared$. Additional operators such as trig functions can be added as needed. 

\subsubsection{Evolution Epochs}
The evolution process relies on a multi-objective fitness function, which utilizes Pareto optimality of correlation and complexity. The Pareto front is then considered to be the models with the best trade-off between correlation and complexity. 

The models evolve using crossover and mutation. The crossover method is a two point crossover modified for the stack data structure. It works by selecting two points in the operator stack and swapping the operators and associated variables/constants between the two points with a similar section in another model.  

The mutation method allows for 6 different types of mutation: variable/constant point mutation, math operator point mutation, pushing new variables and operators to the top of the stacks, trimming off the bottom of the stack, pushing new variables and constants to the bottom of the stacks, and insertion of new operators at a random position in the stack.

The goal population size is 300 models. For each generation 79\% of the 300 models are generated by mutating models from the previous generation. 11\% of the 300 models are generated using crossover on model pairs from the previous generations. 10\% of the 300 models are randomly generated. Additionally, 10\% of models nearest the Pareto front of the previous generation are cloned into the new generation. This leads to populations that generally have between 300 and 330 models.  The models selected for mutation and crossover are chosen via Pareto tournament selection. 

Pareto tournament selection \cite{tournament} works by randomly selecting a subset of models from populations and returning the Pareto front of those subsets. The Pareto tournament size was set to 5 models. The small tournament size was chosen to promote diversity while preventing several models from dominating a population. An example tournament with 8 models is shown in Figure 1. 

\begin{figure}[h]
\caption{Example Pareto Tournament Selection: This figure represents a Pareto tournament of size 8 where the models on the Pareto front are highlighted. The highlighted models (larger points) would be considered the winners of the tournament and all of the models on the Pareto front would be returned.}
\label{fig:pareto}
\includegraphics[width=8cm]{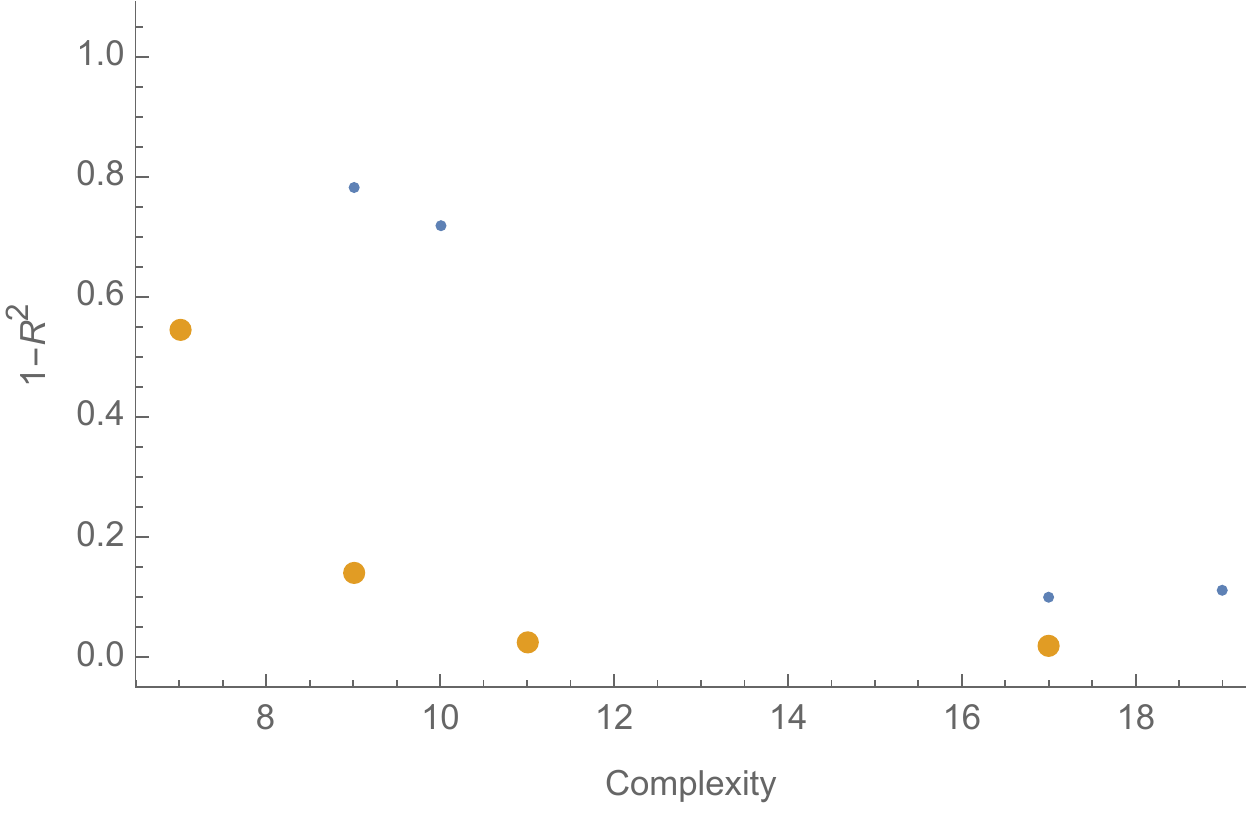}
\end{figure}

\subsubsection{Ensemble Generation and Data Selection}

Once the models are developed, an ensemble is generated. The ensemble is generated by partitioning the current training data using Mathematica's built in clustering capabilities  \cite{clusteringcomponents} and selecting models that best fit each data partition \cite{ensemble_method}\footnote{Mathematica's ClusteringCompenents function is used with default settings where the method for clustering is automatically selected based on what Mathematica determines is the best suited method for the supplied data set.}.Early in the active learning process the ensembles will be smaller since the number of data clusters is limited by the number of data points. As the number of data points increases, the limit on the number of clusters increases. The maximum number of clusters is capped at 10 to ensure ensembles do not grow beyond 10 models. This helps prevent ensemble evaluations from becoming too computationally intense. In the event that all data points are similar and only a single cluster forms, the Pareto front of the models is chosen as the ensemble rather than a single model. 

Using the selected ensemble, Mathematica's NMaximize\footnote{Mathematica's NMaximize function was used with default settings which allows Mathematica to choose a maximization method from the following options: Nelder Mead, Differential Evolution, Simulated Annealing, and Random Search.} \cite{nmaximize} function is employed to find the data point that maximizes the uncertainty metric defined in section 3.1, within the bounds of each variable as described in the Feynman Symbolic Regression Database \cite{data}. It is possible that a local maximum is found rather than a global maximum. This is acceptable since the point still represents a point of relatively high uncertainty. The parameters found to maximize uncertainty are then used to collect the true model response. This data is then added to the training set and will be used in the next run of model evolution. 

It is possible a point that already exists in the training set is selected as the new point. If this occurs, rather than duplicating the point, a new point is selected by maximizing the uncertainty of the ensemble in a random region of the original search space. This helps ensure that new information is being added in each iteration. 

Once the new training point has been added to the training set, another evolutionary epoch begins. This new evolutionary epoch is seeded with the 20\% of models nearest the Pareto front from the previous epoch. This ensures that good models are not lost between evolutionary epochs. This does introduce the risk that these more developed models will dominate over the less developed models at the beginning of the new epoch and bias the evolution. This risk is limited by the small tournament sizes, although it could be further limited in the future using other methods such as age layering. 

This learning process is repeated until a perfect model is found or a maximum number of iterations has completed.

\subsubsection{Reporting}

Once the iterations have completed a report is generated. This report contains the best model that was found during the search, error plots of the best model found in each iteration, the training points used, and the total number of training points needed before finding the perfect model (if one was found).

\subsection{Ensemble Design}
The ensembles are created by partitioning the training data using Mathematica's built-in clustering function, ClusteringComponents. The data set is partitioned into a max of 10 clusters. All of the models are then evaluated over each partition and their average error for each partition is computed. The model with least error for each partition is selected and added to the ensemble, ensuring no model is chosen more than once. If a model has already been selected for a different partition the next best model is selected. This helps ensure ensemble diversity. 

It is possible for the clustering algorithm to fail and return only 1 cluster when given a small data set of all similar points. Since a single model can't act as an ensemble, when this occurs, the Pareto front of the current model set is chosen as the ensemble. 
\subsection{StackGP}
StackGP is the stack based genetic programming system used to evolve models during the model development step of the active learning strategy. StackGP is built for use within Mathematica. The key components of this system are described below in the following subsections. 
\subsubsection{Model Form}
StackGP is the system being used for symbolic regression. Models are represented as stacks, where data types are stored in separate stacks. The operators are stored in the operator stack and the variables/constants are stored in another stack. The evaluation of each model is driven by the operator stack, such that it grabs the next available operator and pops off the variable/constant stack as many variables/constants as the operator needs. This continues until no more operators are left in the operator stack. 

\subsubsection{Genetic Operators}
Models are evolved using three primary genetic operators: mutation, recombination, and cloning. Two point crossover is used as the recombination operator. Mutation has multiple choices for how it will modify the parent models to produce the offspring and these choices are randomly chosen each time the operator is called. The types of mutation are as follows: variable/constant point mutation, math operator point mutation, pushing new variables and operators to the top of the stacks, trimming off the bottom of the stack, pushing new variables and constants to the bottom of the stacks, and insertion of new operators at a random position in the stack.

\subsubsection{Selection/Fitness}
The models compete via Pareto tournaments, wherein each tournament 5 models are randomly selected and the Pareto front of the models are returned as the winners. The Pareto front consists of all the models that are not dominated in either accuracy or simplicity. Correlation is used as the accuracy metric, and combined stack length is used as the complexity metric. The selected models are then added to the pool of models that are assigned to be either cloned, mutated, or paired with another model from the pool of selected models for recombination. 

\subsubsection{Termination Criteria}
Models are evolved until either a set number of generations has completed or until a time limit has been reached. The termination criteria was set to be 2 minutes with no limit on the number of generations. 

\subsubsection{Default Evolution \& Active Learning Parameters}
The parameters for evolution were selected using an active learning strategy for parameter optimization. The optimization goal was to reduce the number of data points needed to find correct models on a sample problem. The parameters that were optimized were the following: mutation rate, crossover rate, spawn rate, elitism rate, tournament size, population size, selection rate, and crossover method. The training set was initialized with results from testing 3 different parameter settings on the sample problem. The active learning for parameter selection began by training models to fit those 3 parameter settings to the number of points needed to solve the sample problem. The developed models were then used to predict the best configuration and then tests were performed using the predicted parameter settings. The results of using the recommended parameter settings were then returned to the training set. This process was iterated until convergent behaviour was observed. Table 1 shows the resulting parameter settings.  The mutation rate shown in Table 1 refers to the rate at which a model will receive exactly one mutation.

\begin{table}
\caption{StackGP \& Active learning Parameter Settings}
\label{tab:parameters}
\begin{tabular}{lc} 
 \toprule
Parameter & Setting\\
\midrule
 Mutation Rate & 79 \\ 
 Crossover Rate & 11 \\
 
 Spawn Rate & 10\\
 
 Elitism Rate & 10\\
 
 Crossover Method & 2 Pt.\\
 
 Tournament Size & 5\\
 
 Population Size & 300\\
 
 Selection Rate & 20\\
\bottomrule
\end{tabular}
\end{table}

\subsection{The Feynman Symbolic Regression Benchmark}

The Feynman set of equations consists of many physics equations compiled by Richard Feynman. A recent paper by Udrescu and Tegmark used the Feynman set of equation as inspiration for their physics based symbolic regression system\cite{feynman}. As well, they selected a subset of 100 Feynman equations that don't use derivatives or integrals and created the Feynman Symbolic Regression Database\cite{data}.  In their paper they reported the minimum number of data points needed by their system, AIFeynman, to solve each physics problem. They also reported the noise tolerance, whether their system had to use more sophisticated techniques to solve each equation, the runtime to solution, and whether another industry grade symbolic regression system, Eureqa, was able to solve it. The results in that paper and that data set allow for direct comparisons of the success of new systems to both AIFeynman and Eureqa.

\section{Results}

37 of the 100 equations we were able to solve with just the initial random 3 data points. The minimum number of points needed by AIFeynman was 10, so StackGP outperformed AIFeynman on all of these problems. This indicates that these problems are trivially solvable and active learning is not necessary for these problems, so they give no insight into how active learning is affecting the search. Of the remaining problems, 16 were solved using fewer data points than what was reported by AIFeynman. For these problems, it seems that the active learning had a positive effect on the success of the search. One of the equations needed the same number of points as AIFeynman. 18 of the problems required more data points than what was reported by AIFeynman. The 28 remaining problems were not solved within 100 iterations of active learning, so it is not possible to compare the effect that active learning had on the success of those searches. The results are summarized in Table 2. 

According to Udrescu and Tegmark, Eureqa is the best available commercial symbolic regression software \cite{feynman}. Eureqa was found to solve 71 of the 100 Feynman equations using 300 data points and 2 hours of compute time for each equation. StackGP with active learning was able to find 72 of the 100 Feynman equations, so performed similarly to Eureqa, although not all the same equations were solved. 

\begin{table}
\caption{StackGP with Active Learning Performance Summary}
\label{tab:summary}
\begin{tabular}{lc} 
 \toprule
 Performance & Total Equations\\ 
 \midrule
 Trivial (Only 3 Points Needed) & 37 \\ 
 
 Outperformed AIFeynman & 16 \\
 
 Underperformed AIFeynman & 18\\

 Matched AIFeynman & 1\\
 
 Failed to Solve & 28\\
 \bottomrule
\end{tabular}
\end{table}

The performance on each individual equation is shown in Tables 4 and 5 in the Appendix. The formulae for each equation number alongside the variable ranges and sample data can be found in the Feynman Symbolic Regression Database \cite{data} where they are ordered in the same way. The table shows the number of data points needed to solve each equation by AIFeynman, the number of data points needed to solve each equation by StackGP with active learning, the success of StackGP with active learning, and the success of Eureqa. The number in parenthesis indicates the number of repeated trials completed and averaged (median) to get the total number of points needed to solve the problem. Many of the equations were able to be tested using 100 repeated trials, although some were tested fewer times due to limited access to computing resources. 

In the following we discuss a few examples.
Equation number 22 is an example of a problem that needed just 3 points to be found. 

\begin{equation} \tag{Eq 22}
\tau=r F  \sin(\theta) 
\end{equation}

Looking at the equation we can see that it is relatively simple and would require only 3 operators (sin, *, *) and 3 variables (r, f, $\theta$). It is likely easy to find, both due to its simplicity and since the terms are combined as products, which makes each variables contribution to the response data easily distinguishable and similar in magnitude.

Equation number 3 is an example of an equation where the active learning approach with StackGP outperformed AIFeynman, needing just 42.5 points on average compared to the 1000 points needed by AIFeynman. As well, this specific equation was unsolvable by Eureqa. 

\begin{equation}\tag{Eq 3}
f=\frac{e^{-\frac{1}{2} \left(\frac{\theta -\theta _1}{\sigma }\right){}^2}}{\sqrt{(2 \pi)}\sigma}
\end{equation}

Equation number 5 is an example of an equation that was unsolvable by StackGP with active learning and by Eureqa. 
\begin{equation}\tag{Eq 5}
F=\frac{G m_1 m_2}{\left(x_2-x_1\right){}^2+\left(y_2-y_1\right){}^2+\left(z_2-z_1\right){}^2} 
\end{equation}

\noindent
It required 1 million data points to be solved by AIFeynman. This equation is rather complicated since it has 9 variables and the contributions of each variable to the response are vastly different depending on where they are in the equation.

\section{Ablation Study}

An ablation study was completed to determine if the performance of StackGP using active learning can be attributed to the active learning strategy or the stack based genetic programming system. The ablation study compared the active learning approach against a modified approach where each new data point was randomly selected rather than selected according to the active learning strategy. A sample of equations from the set were chosen to highlight a variety of equation forms. Specifically, equations 1, 2, and 3 were chosen to highlight how very similar but slightly modified equations can differ in how successful this active learning approach is. 

It is unexpected that active learning works as well as it does on equation 3, when it performs worse than random point selection on equation 2, despite equation 2 being similar yet simpler than equation 3. For comparison, equation 1, 2, and 3 are shown below. A variable is added between each equation making equation 2 slightly more complex than equation 1 and equation 3 slightly more complex than equation 2.  The expected behaviour would be for the equations to be more difficult to find as the complexity increases. When using random point selection, this behaviour is observed, but when using active learning, equation 2 seems to be more difficult to find than equation 3. It is unclear what would make equation 2 more difficult for active learning than when random point selection is used, so further analysis of equation 2 could be useful. 
\begin{equation}\tag{Eq 1}
f =    \frac{e^{-\frac{\theta ^2}{2}}}{\sqrt(2 \pi )}
\end{equation}

\begin{equation}\tag{Eq 2}
f = \frac{e^{-\frac{1}{2} \left(\frac{\theta }{\sigma }\right)^2}}{\sqrt{2 \pi } \sigma }    
\end{equation}

\begin{equation}\tag{Eq 3}
f =   \frac{e^{-\frac{1}{2} \left(\frac{\theta -\theta _1}{\sigma }\right){}^2}}{\sqrt{(2 \pi ) }\sigma }
\end{equation}

Equation 24 is another example of an equation that active learning performed well on and excelled over random point selection. 
\begin{equation}\tag{Eq 24}
 E =    \frac{1}{4} m x^2 \left(\omega ^2+\text{$\omega $1}^2\right) 
\end{equation}

Equation 14 and Equation 47 both showed worse performance using active learning over random search. It is possible that the active learning point selection is mislead by these equations to select points that are not maximally informative or points that are very similar to points previously selected. 

\begin{equation}\tag{Eq 14}
 U =  G m_1 m_2 \left(\frac{1}{r_2}-\frac{1}{r_1}\right) 
\end{equation}

\begin{equation}\tag{Eq 47}
\kappa =   \frac{k v}{A (\gamma -1)}  
\end{equation}

\begin{table}
\caption{Average number of points needed in Active Learning (AL) vs. Random Point Selection (Random). Number of trials for reference.}
\label{tab:ablation}
\begin{tabular}{cccl} 
 \toprule
 EQ\# & AL & Random & Trials\\ 
 \midrule
 1 & 3 & 3 & 100\\ 
 
 2 & 43 & 28.5 & 100\\
 
 3 & 42.5 & $>$202 & 100\\
 
 4 & 25 & 26 & 100 \\
 
 10 & 4 & 5 & 100 \\
 
 11 & 3 & 3 & 100 \\
 
 12 & 3 & 3 & 100 \\
 
 14 & 19.5 & 14 & 100 \\
 
 15 & 3 & 3 & 100 \\
 
 16 & 3 & 3 & 100 \\
 
 23 & 4 & 4 & 100 \\
 
 24 & 28.5 & 49.5 & 100 \\
 
 32 & 10.5 & 11 & 100 \\
 
 47 & 28.5 & 17.5 & 40 \\
 
 60 & 8 & 7.5 & 100 \\
 
 61 & 30 & 30.5 & 40 \\ 
 \bottomrule
\end{tabular}
\end{table}

\section{Discussion}

Although the active learning approach did not outperform AIFeynman on all of the tested equations, it does show promise in that it has a similar success rate to Eureqa and uses no domain expertise, unlike AIFeynman. The active learning approach also represents a self-guided experimentation process where the machine learning algorithm can direct an experimentation and design process so that researchers can spend less time planning their next experiments. Even more important, with active learning, it is less likely that an exhaustive set of experiments needs to be completed to fully understand a system of study. 

It was observed in the ablation study that some types of problems are better suited for this method of active learning while other types of problems are more difficult for this active learning method than random point selection. For those problems that are difficult for this active learning approach it is possible that the active learning point selection is being misled to choose points that are not actually maximally informative. An attempt to avoid this problem was made by introducing the random subspace point selection, since previously an issue would occur for some problems where identical points were chosen many times over. However, this approach was likely not sufficient to ensure similar points are not repeatedly selected. Further work will explore an approach where new points have to be a minimum distance away from points already in the data set to ensure that similar points are not gathered at every selection event. Alternatively, a hybrid random and informed point selection could be utilized.

A second possibility is that the issue lies with the uncertainty metric used. The uncertainty metric used is a {\it relative} measure since it is scaled by the magnitude of the ensemble response. This seemed like a good approach since the magnitude of uncertainty would likely increase as the magnitude of the response increases, but that may not always be the case. If it is not the case, point selection could become biased towards regions where the uncertainty metric is magnified by the smaller ensemble response. This could potentially be fixed by changing the uncertainty metric to not be relative to the magnitude of the ensemble response. 

Further research to explore how well various uncertainty metrics affect the success of this active learning approach on various problem types will be useful. As well, it could be beneficial to explore and classify the types of problems that tend to be difficult or easy for this active learning method. 

From observation it seems that of all the equations in the Feynman data set, the ones that tend to pose difficulty for this active learning approach tend to have complex denominators. This could support the concern that the relative uncertainty metric is being mislead for some problem types. Alternatively, it could highlight a weakness with current symbolic regression implementations since Eureqa struggled with many of those problems as well. It is possible that when variables exist in the denominators of problems, that it becomes more difficult for symbolic regression to determine the true contribution of those variables. To determine if this is a larger scale weakness with symbolic regression it could be useful to compare several additional symbolic regression implementations on those difficult problems.  

Future research is planned to explore the applications of other active learning techniques, both model and data driven, to genetic programming with the goal of determining which methods are most successful for different classifications of problems. This information will hopefully act as a basis for an active learning toolkit that can be used by researchers in various fields to accelerate their data collection process.

\begin{acks}
Left out from anonymous submission
\end{acks}

\bibliographystyle{ACM-Reference-Format}
\bibliography{biblio}

\newpage

\section{Appendices}

\subsection{Additional Data}
Table~\ref{tab:app1} and Table ~\ref{tab:app2} show the full results for all 100 of the Feynman symbolic regression problems.
\begin{table}
\caption{Number of Data Points Needed to Solve Problems 1-50. For The StackGP solution the number of points is the median of points used out of the indicated number of trials. For Eureqa, 300 points were used for each equation with a 2 hour time limit. }
\label{tab:app1}
\begin{tabular}{ccccl} 
 \toprule
 EQ  & AIFeynman & StackGP Pts & StackGP  & Eureqa \\ [0.5ex]
 Num & Data Pts  & (num trials) & Success & Success\\ [0.5ex]
 \midrule
 1 & 10 & 3 (100) & Yes & No \\ 
 \hline
 2 & 100 & 43 (100) & Yes & No\\
 \hline
 3 & 1000 & 42.5 (100) & Yes & No \\
 \hline
 4 & 100 & 25 (100) & Yes & No \\
 \hline
 5 & 1000000 & - (100) & No & No \\
 \hline
 6 & 10 & - (100) & No & No \\
 \hline
 7 & 100 & $>$102 (100) & No & Yes \\
 \hline
 8 & 10 & 3 (100)  & Yes & Yes\\
 \hline
 9 & 10 & 68 (1)  & Yes & Yes\\
 \hline
 10 & 10 & 4 (100)  & Yes & Yes\\
 \hline
 11 & 10 & 3 (100)  & Yes & Yes\\
 \hline
 12 & 10 & 3 (100)  & Yes & Yes\\
 \hline
 13 & 10 & 21 (1)  & Yes & Yes\\
 \hline
 14 & 10 & 19.5 (100)  & Yes & Yes\\
 \hline
 15 & 10 & 3 (100)  & Yes & Yes\\
 \hline
 16 & 10 & 3 (100)  & Yes & Yes\\
 \hline
 17 & 10 & -  & No & No\\
 \hline
 18 & 100 & -  & No & No\\
 \hline
 19 & 10 & -  & No & No\\
 \hline
 20 & 10 & -  & No & No\\
 \hline
 21 & 10 & -  & No & Yes\\
 \hline
 22 & 10 & 3 (100)  & Yes & Yes\\
 \hline
 23 & 10 & 4 (100)  & Yes & Yes\\
 \hline
 24 & 10 & 28.5 (100)  & Yes & Yes\\
 \hline
 25 & 10 & 3 (100)  & Yes & Yes\\
 \hline
 26 & 100 & 3 (100)  & Yes & Yes\\
 \hline
 27 & 10 & 12 (100)  & Yes & Yes\\
 \hline
 28 & 10 & 3 (100)  & Yes & Yes\\
 \hline
 29 & 1000 & -  & No & No\\
 \hline
 30 & 100 & -  & No & Yes\\
 \hline
 31 & 100 & 3 (1)  & Yes & Yes\\
 \hline
 32 & 10 & 10.5 (100)  & Yes & Yes\\
 \hline
 33 & 10 & -(100)  & No & No\\
 \hline
 34 & 10 & 3 (1)  & Yes & Yes\\
 \hline
 35 & 10 & 4 (1)  & Yes & No\\
 \hline
 36 & 10 & -  & No & No\\
 \hline
 37 & 10 & 3 (1)  & Yes & Yes\\
 \hline
 38 & 100 & -  & No & Yes\\
 \hline
 39 & 10 & 11 (1)  & Yes & Yes\\
 \hline
 40 & 10 & 3 (1)  & Yes & Yes\\
 \hline
 41 & 10 & 9 (1)  & Yes & Yes\\
 \hline
 42 & 10 & 3 (1)  & Yes & Yes\\
 \hline
 43 & 10 & 102 (1)  & Yes  & No\\
 \hline
 44 & 10 & -  & No & No\\
 \hline
 45 & 10 & 3 (1)  & Yes & Yes\\
 \hline
 46 & 10 & 3 (1)  & Yes & Yes\\
 \hline
 47 & 10 & 28.5 (40)  & Yes & Yes\\
 \hline
 48 & 10 & 32 (1)  & Yes & Yes\\
 \hline
 49 & 10 & 3 (1)  & Yes & Yes\\
 \hline
 50 & 100 & -  & No & No\\ 
  \bottomrule
\end{tabular}
\end{table}

\newpage

\begin{table}
\caption{Number of Data Points Needed to Solve Problems 51-100. For The StackGP solution the number of points is the median of points used out of the indicated number of trials. For Eureqa, 300 points were used for each equation with a 2 hour time limit.}
\label{tab:app2}
\begin{tabular}{ccccl} 
 \toprule
 EQ & AIFeynman & StackGP Pts   & StackGP  & Eureqa \\ [0.5ex]
 Num & Data Pts & (num trials) & Success & Success\\ [0.5ex]
 \midrule
 51 & 10 & -  & No & Yes\\
 \hline
 52 & 10 & 13 (1)  & Yes & Yes\\
 \hline
 53 & 10 & 3 (1)  & Yes & Yes\\
 \hline
 54 & 10 & 3 (1)  & Yes & Yes\\
 \hline
 55 & 10 & 13 (1)  & Yes & Yes\\
 \hline
 56 & 10000 & -  & No & No\\
 \hline
 57 & 10 & 92 (1)  & Yes & Yes\\
 \hline
 58 & 10 & 3 (1)  & Yes & Yes\\
 \hline
 59 & 10 & 3 (1)  & Yes & Yes\\
 \hline
 60 & 10 & 8 (100)  & Yes & Yes\\
 \hline
 61 & 10 & 30 (100)  & Yes & Yes\\
 \hline
 62 & 10 & 76.5 (100) & Yes & Yes\\
 \hline
 63 & 10 & 14 (1)  & Yes & Yes\\
 \hline
 64 & 100 & -  & No & No\\
 \hline
 65 & 100 & -  & No & No\\
 \hline
 66 & 10 & 5 (100)  & Yes & Yes\\
 \hline
 67 & 100 & 68 (1)  & Yes & No\\
 \hline
 68 & 10 & -  & No & No\\
 \hline
 69 & 10 & 3 (1)  & Yes & Yes\\
 \hline
 70 & 10 & 3 (1)  & Yes & Yes\\
 \hline
 71 & 10 & 22 (1)  & Yes & Yes\\
 \hline
 72 & 10 & -  & No & No\\
 \hline
 73 & 10 & 3 (100)  & Yes & Yes\\
 \hline
 74 & 10 & 3 (100) & Yes & Yes\\
 \hline
 75 & 10 & 3 (100)  & Yes & Yes\\
 \hline
 76 & 10 & 3 (100)  & Yes & Yes\\
 \hline
 77 & 10 & 3 (100)  & Yes & Yes\\
 \hline
 78 & 10 & 3 (100)  & Yes & Yes\\
 \hline
 79 & 10 & 3 (1)  & Yes & Yes\\
 \hline
 80 & 10 & -  & No & No\\
 \hline
 81 & 10 & -  & No & Yes\\
 \hline
 82 & 10 & -  & No & Yes\\
 \hline
 83 & 10 & 4 (100)  & Yes & Yes\\
 \hline
 84 & 10 & 3 (100)  & Yes & Yes\\
 \hline
 85 & 10 & 4 (1)  & Yes & Yes\\
 \hline
 86 & 10 & -  & No & No\\
 \hline
 87 & 10 & -  & No & No\\
 \hline
 88 & 10 & 3 (1)  & Yes & Yes\\
 \hline
 89 & 10 & 6 (1)  & Yes & No\\
 \hline
 90 & 1000 & -  &  & No\\
 \hline
 91 & 100 & 83 (1)  & Yes & Yes\\
 \hline
 92 & 10 & 3 (1)  & Yes & Yes\\
 \hline
 93 & 10 & 5 (1)  & Yes & Yes\\
 \hline
 94 & 10 & -  & No & No\\
 \hline
 95 & 10 & 10 (1)  & Yes & Yes\\
 \hline
 96 & 10 & 3 (100)  & Yes & Yes\\
 \hline
 97 & 10 & 3 (100)  & Yes & Yes\\
 \hline
 98 & 10 & 7 (1)  & Yes & Yes\\
 \hline
 99 & 10 & 9 (1)  & Yes & Yes\\
 \hline
 100 & 10 & 3 (100)  & Yes & Yes\\ 
  \bottomrule
\end{tabular}
\end{table}



\end{document}